\documentclass[final,12pt]{elsarticle}
\usepackage{hyperref}
\usepackage{tabularray}
\usepackage{graphicx}
\usepackage{dsfont}
\usepackage{indentfirst}
\usepackage{amsthm,amsmath,amssymb}
\usepackage{dutchcal}
\usepackage{makecell}
\usepackage[labelfont=bf]{caption}
\usepackage{booktabs}
\usepackage{epsfig}
\usepackage{subcaption}
\usepackage{caption}
\usepackage{floatrow}
\usepackage[ruled]{algorithm2e}
\newfloatcommand{capbtabbox}{table}[][\FBwidth]
\usepackage[final]{changes}

\makeatletter
\def\ps@pprintTitle{%
 \let\@oddhead\@empty
 \let\@evenhead\@empty
 \def\@oddfoot{}%
 \let\@evenfoot\@oddfoot}
\makeatother

\journal{Knowledge-Based Systems}

\begin{document}
\begin{frontmatter}

\title{Learning Robust Correlation with Foundation Model for Weakly-Supervised Few-Shot Segmentation}

\author[author1]{Xinyang Huang}
\author[author1]{Chuang Zhu\corref{cor1}}
\author[author1]{Kebin Liu}
\author[author1]{Ruiying Ren}
\author[author1]{Shengjie Liu}

\address[author1]{School of Artificial Intelligence, Beijing University of Posts and Telecommunications, Beijing, China}
\cortext[cor1]{Corresponding author}

\begin{abstract}
Existing few-shot segmentation (FSS) only considers learning support-query correlation and segmenting unseen categories under the precise pixel masks.
However, the cost of a large number of pixel masks during training is expensive.
This paper considers a more challenging scenario, weakly-supervised few-shot segmentation (WS-FSS), which only provides category ($i.e.$ image-level) labels.
It requires the model to learn robust support-query information when the generated mask is inaccurate.
In this work, we design a Correlation Enhancement Network (CORENet) with foundation model, which utilizes multi-information guidance to learn robust correlation.
Specifically, correlation-guided transformer (CGT) utilizes self-supervised ViT tokens to learn robust correlation from both local and global perspectives.
From the perspective of semantic categories, the class-guided module (CGM) guides the model to locate valuable correlations through the pre-trained CLIP.
Finally, the embedding-guided module (EGM) implicitly guides the model to supplement the inevitable information loss during the correlation learning by the original appearance embedding and finally generates the query mask.
Extensive experiments on PASCAL-5$^i$ and COCO-20$^i$ have shown that CORENet exhibits excellent performance compared to existing methods.
Our code will be available soon after acceptance.
\end{abstract}

\end{frontmatter}

\renewcommand{\thefootnote}{\fnsymbol{footnote}}
\footnotetext[0]{Email: hsinyanghuang7@gmail.com; czhu@bupt.edu.cn}
\renewcommand{\thefootnote}{\arabic{footnote}}

\section{Introduction}
\label{introduction}
Few-shot learning \cite{fei2006one,wang2020generalizing,snell2017prototypical,xie2020secure,qin2021prior,zhang2023autonomous} is a machine learning method that uses very little labeled data to help the model quickly adapt to new tasks or categories.
It is crucial in applications where data collection is costly or requires intensive annotation, such as image segmentation. 
Consequently, few-shot segmentation (FSS) has been proposed and extensively studied \cite{shaban2017one,wang2019panet,zhang2019canet, liu2020crnet,tian2020prior,min2021hypercorrelation,yang2023mianet,li2023lite}.

\begin{figure}[t]
	\centering
	\begin{subfigure}{0.295\linewidth}
		\centering
		\includegraphics[width=1\linewidth]{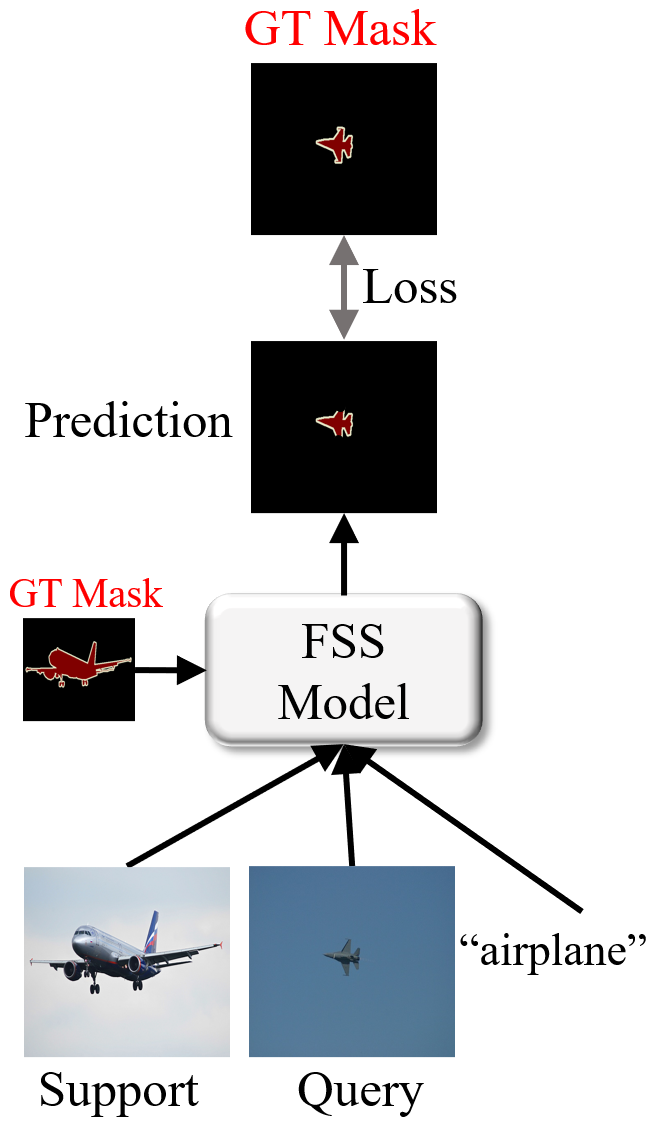}
		\caption{FSS}
		\label{introa}
	\end{subfigure}
	\centering
	\begin{subfigure}{0.295\linewidth}
		\centering
		\includegraphics[width=1\linewidth]{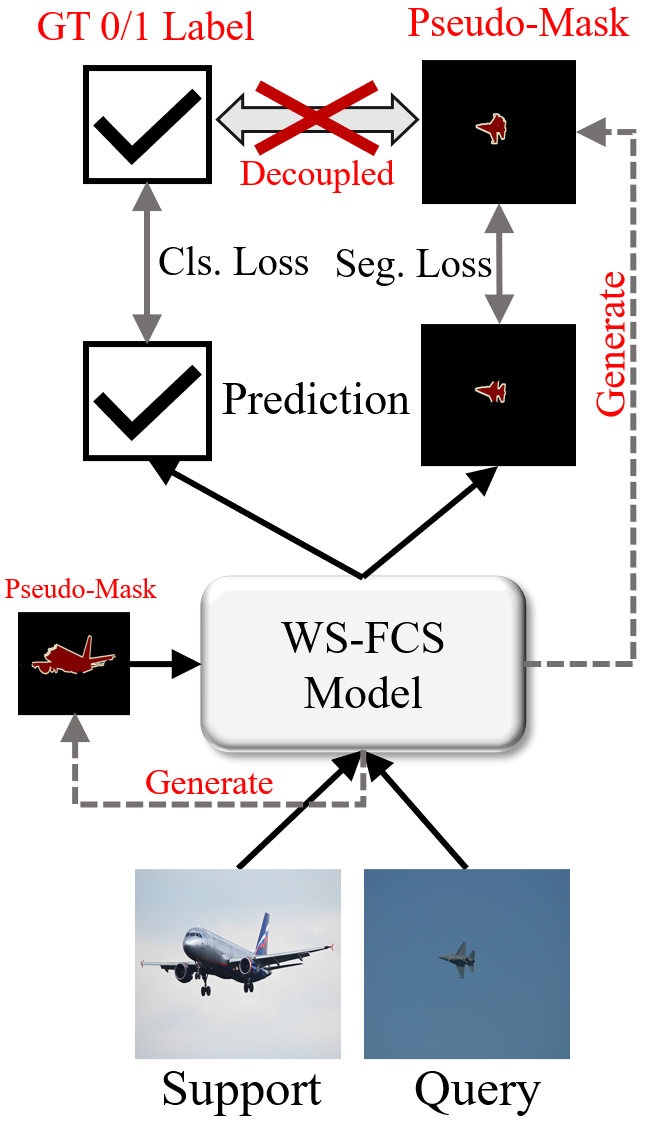}
		\caption{WS-FCS}
		\label{introb}
	\end{subfigure}
	\centering
	\begin{subfigure}{0.33\linewidth}
		\centering
		\includegraphics[width=1\linewidth]{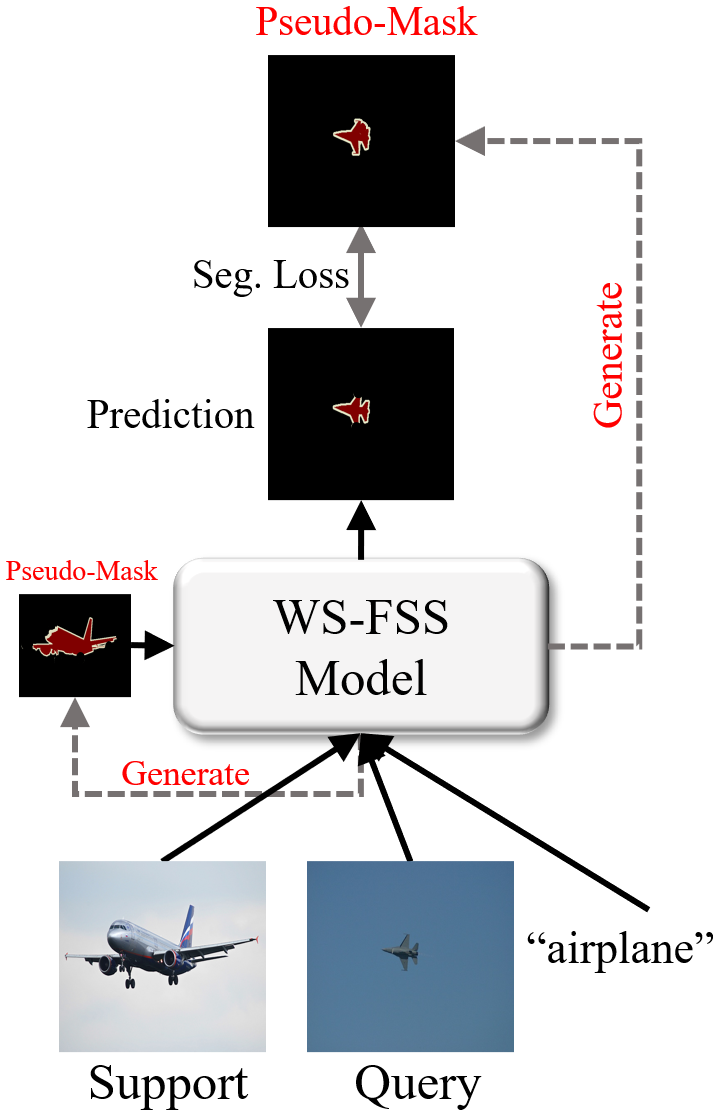}
		\caption{WS-FSS}
		\label{introc}
	\end{subfigure}
\caption{Comparison between (a) few-shot segmentation (FSS) task \cite{shaban2017one}, (b) weakly-supervised few-shot classification and segmentation (WS-FCS) task \cite{kang2023distilling}, and (c) our weakly-supervised few-shot segmentation (WS-FSS) task settings.
(a) The FSS task requires many support-query masks during training.
(b) The classification and segmentation tasks are decoupled in the WS-FCS task.
It provides supervisory information on whether images belong to the same category without providing specific category assistance for segmentation.
(c) The WS-FSS task assists the model in segmentation through specific categories of supervised information in the presence of noise in the mask generated by the model.
}
\end{figure}

Existing FSS methods are typically trained based on the meta-learning paradigm \cite{finn2017model,schmidhuber1987evolutionary,nichol2018reptile,rivolli2022meta,feng2022meta}. 
They often assume the presence of a large amount of accurately annotated data for model training and learn the support-query correlation by abundant support and query masks, as shown in Fig.~\ref{introa}. 
Likewise, during testing, several ground-truth (GT) support masks are required during the reference of the model. 
However, the cost of obtaining the segmentation masks required for these models is very expensive and cumbersome.

Although related works \cite{gama2022weakly,zhang2022weakly} have explored the setting of few-shot segmentation in weakly-supervised scenarios, they are unable to generate supervised masks for unseen categories during the testing phase or require additional training in a mask generation module.
CST \cite{kang2023distilling} solved this problem and proposed the weakly-supervised few-shot classification and segmentation (WS-FCS), as shown in Fig.~\ref{introb}.
However, there are two issues when directly applying it to WS-FSS tasks: firstly, it simply considers the correlation information of the support-query pair in the presence of the GT mask, which may introduce a lot of matching noise in the case of inaccurate masks; secondly, the provided category information is whether the two images belong to the same category, ignoring the benefits of semantic level information for segmentation.
\cite{lee2022pixel} is closest to the problem setting of this paper, but it ignores the exploration of robust correlation and the contribution and role of the foundation model in the WS-FSS task.
Similar to \cite{lee2022pixel}, this paper focuses on a weakly-supervised few-shot segmentation (WS-FSS) scenario where the model should learn robust support-query matching information and perform segmentation on query images \textbf{with only image-level category information and no access to GT masks}, as depicted in Fig.~\ref{introc}. 

To solve the WS-FSS, this paper introduces a \textbf{Correlation Enhancement Network} (CORENet) with foundation model assistance that helps the model learn robust correlation from multiple perspectives, even in the presence of inaccurate masks generated by the model.
Specifically, we first design a Correlation-Guided Transformer (CGT), which takes high-quality tokens obtained from a self-supervised Vision Transformer (ViT) \cite{caron2021emerging} as input. 
It fuses information from local and global perspectives to guide the model in better utilizing correlation information. 
Therefore, CGT can be relatively robust in the face of generated imprecise masks.
Furthermore, a well-designed self-distillation loss helps CGT generate higher-quality correlation maps in the early stages.
However, when the model generates inaccurate masks, the effect of segmenting the query from the perspective of correlation is limited.
To address the above issues, the Class-Guided Module (CGM) helps the model to roughly locate specific objects from inaccurate masks using prior knowledge by using the provided class information.
Although existing works \cite{kang2022integrative,kang2023distilling} utilize category supervision information by classifying support and query during FSS. 
They provide category information to support and query whether the images belong to the same category (0/1 label) without providing specific category semantic information assistance for segmentation.
By using pre-trained CLIP \cite{radford2021learning} to generate coarse attention, CGM utilizes existing correlation features to filter out background features unrelated to the query foreground, helping the model roughly locate valuable correlation information.
Finally, to further reduce potential information loss during correlation processing and implicitly guide the model in refining matching information, we propose an Embedding-Guided Module (EGM). 
EGM uses efficient tokens generated by ViT to supplement the information of the original embeddings, resulting in the final masks.

To generate supervised masks, inspired by previous works \cite{caron2021emerging,kang2023distilling}, the paper utilizes attention maps generated by pre-trained self-supervised ViT to create pseudo-masks. 
Furthermore, we leverage pixel relationships within the image to generate more accurate pseudo-masks through the Pixel-Adaptive Refinement (PAR) module \cite{ru2022learning}, which helps the model learn robust correlations from the perspective of mask enhancement.
Even when encountering unseen categories during testing, the model can provide relatively accurate pseudo-masks.
Our main contributions are summarized as follows:

\begin{itemize}
\item We propose a Correlation Enhancement Network (CORENet) with foundation model assistance to guide models from multiple perspectives to learn robust correlation in WS-FSS.

\item We propose a Correlation-Guided Transformer (CGT) that learns to support-query knowledge from a knowledge aggregation perspective and apply the Pixel Adaptive Refinement (PAR) module in a few-shot scenario for the first time.

\item We propose a Class-Guided Module (CGM) and an Embedding-Guided Module (EGM) to mine and supplement target information in correlation features from category semantics and appearance embedding perspective.

\item Our CORENet achieved state-of-the-art results compared to the latest FSS and WS-FSS methods in two WS-FSS scenarios ($i.e.$ PASCAL-5$^i$ and COCO-20$^i$).
\end{itemize}

The remainder of this paper is as follows:
Section \ref{related work} reviews recent work related to WS-FSS.
Sections \ref{problem deif} and \ref{methodogy} elaborate on the entire process of our proposed CORENet.
Then, comprehensive quantitative and qualitative results are reported in Section \ref{experiments}, followed by a series of ablation studies.
Finally, Section \ref{conclusion} gives the conclusion of this work.

\section{Related Work}
\label{related work}
\textbf{Few-Shot Semantic Segmentation.}
Few-shot semantic segmentation (FSS) aims to segment new semantic objects in images, with only a few densely labeled examples available.
The current methods mainly focus on the improvement of the meta-learning stage.
They can be classified as prototype-based methods and relational-based methods.
The intuition behind the prototype-based methods \cite{siam2019amp,wang2019panet,zhang2019canet,yang2021mining,li2021adaptive,okazawa2022interclass,li2023lite} is to extract representative foreground or background prototypes from the supporting samples using the method, and then use different strategies to interact between different prototypes or between prototypes and query features.
Relational-based methods \cite{liu2020crnet,tian2020prior,min2021hypercorrelation,kang2022integrative,kang2023distilling,yang2023mianet} have also achieved great success in the few-shot semantic segmentation.
However, these methods only focus on learning to support and query matching information between images under precise supervision.
This paper considers a more challenging weak supervision version of FSS, which completes the segmentation of query images without providing any mask information, only providing support images and category information.

\textbf{Weakly-Supervised Few-Shot Segmentation.}
Due to the severe challenge of data scarcity, many works currently study few-shot segmentation in a weakly-supervised environment. 
However, the definition of weakly supervised few-shot segmentation (WS-FSS) in existing methods is still flawed and inconsistent.
WS Co-FCN \cite{raza2019weakly} generated a pseudo-mask to support the image by retaining pixels not classified as background. 
However, it cannot handle supporting images that contain multiple new classes. 
Some methods \cite{gama2022weakly,saha2022improving} use supervision information such as bounding boxes. 
WRCAM \cite{zhang2022weakly} requires pre-training of a mask generation module for all image categories in advance, including test image categories that have not been seen during the training phase, which does not follow the training paradigm of few-shot learning during the training stage. 
The problem setting of CST \cite{kang2023distilling} is similar to that of this paper.
However, the provided category information is whether the two images belong to the same category and does not provide specific category assistance for segmentation.
\cite{lee2022pixel} is closest to the problem setting of this paper, but this paper focuses on exploring the contribution and role of the foundation model in the WS-FSS task.
This paper focuses on the few-shot segmentation in a weakly-supervised scenario, where no GT mask information is provided at any stage. 
It provides category information assistance to complete the segmentation of the query image.

\section{Problem Definition}
\label{problem deif}
Similar to the few-shot segmentation \cite{shaban2017one,wang2019panet,zhang2019canet, liu2020crnet,tian2020prior,min2021hypercorrelation,yang2023mianet}, in order to avoid overfitting risks caused by insufficient training data, we adopted a widely used meta-learning method called episodic training \cite{vinyals2016matching}.
In weakly-supervised few-shot segmentation, we define two datasets, $D_{train}$ and $D_{test}$, with category sets $C_{train}$ and $C_{test}$ respectively, where $C_{train} \cap C_{test}=\varnothing$. 
The model trained on $D_{train}$ is directly transferred to $D_{test}$ for evaluation and testing.
We train the model in an episode manner \cite{vinyals2016matching}.
Under the weak-supervised setting, each episode only comprises support set $S=\{I_s\}$, query set $Q=\{I_q\}$, and their corresponding category $c$.
Unlike few-shot segmentation, we do not provide mask information at any stage.
Under the $K$-shot setting, it includes the support set $S=\{I_s^i\}^K_{i=1}$, query set $Q=\{I_q\}$ and the corresponding category $c$.
Training set $D_{train}$ and test set $D_{test}$ means $D_{train}=\{I_s^i, I_q^i, c\}^{N_{train}}_{i=1}$ and $D_{test}=\{I_s^i, I_q^i, c\}^{N_{test}}_{i=1}$, where $N_{train}$ and $N_{test}$ is a series of quantitative training and testing.
During training, the model iteratively samples an episode from $D_{train}$ to generate a pseudo-mask using limited information and to learn segmentation knowledge through the generated pseudo-masks.
During the testing, the model changed from $D_{test}$ randomly samples $\{I_s^i, I_q^i, c\}$ to predict the query mask.

\begin{figure}[t]
\centering
\includegraphics[scale=0.315]{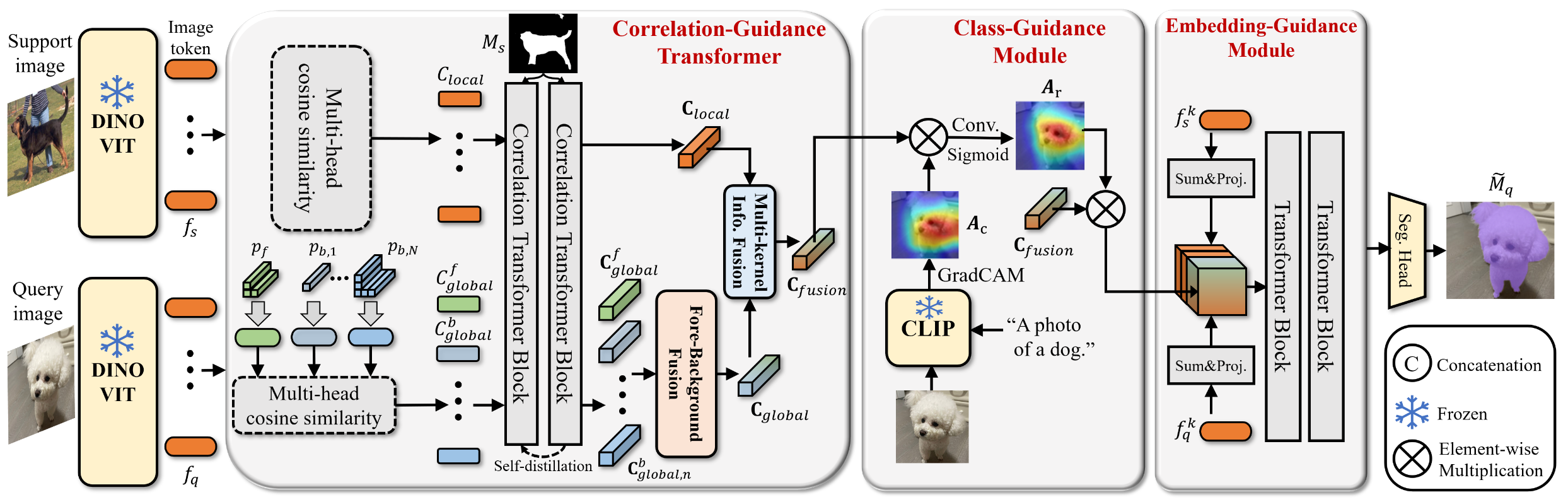}
\caption{The overall architecture of our Correlation Enhancement Network (CORENet).
Firstly, the \textbf{Correlation-Guided Transformer (CGT)} is introduced to generate robust correlation features using the local and global similarity calculations of ViT tokens.
Then, with the assistance of CLIP, the \textbf{Class-Guided Module (CGM)} transforms the category information into coarse attention and further refines them to filter irrelevant information in the relevant features.
Meanwhile, the \textbf{Embedding-Guided Module (EGM)} combines the support query appearance of each layer with the enhanced correlation features, further reducing the potential information loss of the model in correlation-enhanced learning under weakly-supervised settings and obtaining the final query mask.}
\label{overview}
\end{figure}

\section{Methodology}
\label{methodogy}
\subsection{Overview}
As shown in Fig.~\ref{overview}, the Correlation Enhancement Network (CORENet) is composed of three key modules, namely, correlation-guided transformer (CGT), class-guided module (CGM), and embedded-guided module (EGM).
Precisely, we extract high-quality features through pretrained DINO ViT \cite{caron2021emerging} and calculate the correlation between the token of the supporting image pair and the query image pair.
Then, the robust cross-correlation information is learned from the local and global perspectives through CGT. 
With the assistance of CLIP \cite{radford2021learning}, CGM uses category information to guide the generation of a coarse attention map and filters out the irrelevant information in the query features through the generated cross-correlation features.
To reduce the potential information loss of the model in correlation reinforcement, we propose EGM, which further aggregates the matching information by using the embedded information obtained from the feature graph to guide the model to learn the matching information implicitly.
Then, the model sends the learned robust features into the segmentation header to predict the final segmentation mask $\tilde{M}_q$ of the query image.
Next, each module will be described in detail in the following paragraphs.

\subsection{Correlation-Guided Transformer}
\label{sec:cgt}
The correlation between support and query plays a crucial role in FSS.
The existing methods \cite{min2021hypercorrelation, kang2022integrative, kang2023distilling} help the model segment the query image on the existing support foreground information by using the similarity between the support and query image pixels.
However, due to the lack of a GT mask, it is not comprehensive to only consider the correlation information of this local-to-local matching.
In this paper, the correlation-guided transformer (CGT) is proposed. 
From the perspective of local-to-local and local-to-global, CGT uses the features extracted by self-supervised pretrained ViT to learn the multi-view robust correlation information between support images and query images.

\begin{figure}[t]
\centering
\includegraphics[scale=0.4]{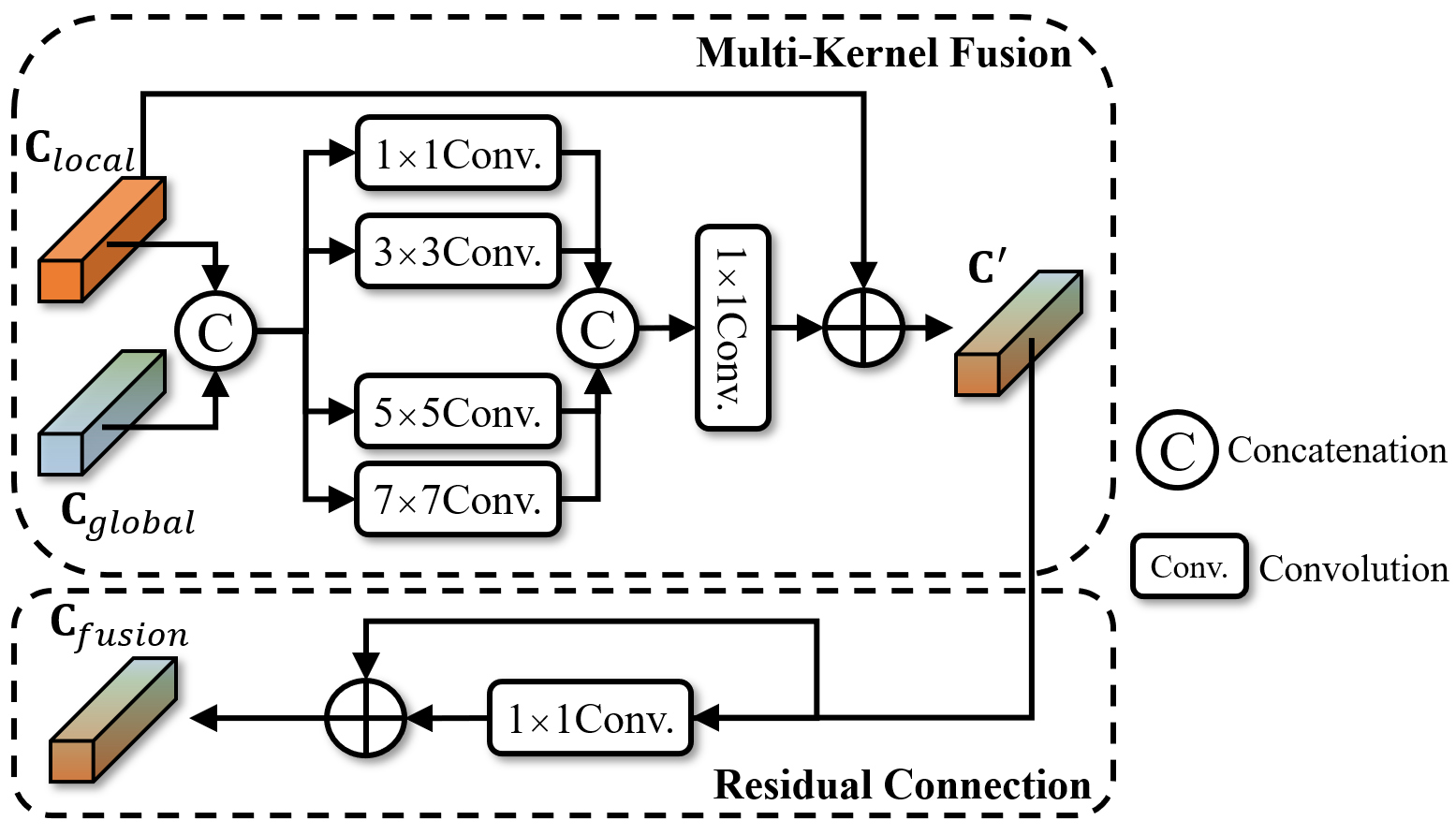}
\caption{Illustration of Multi-kernel information fusion in CGT.}
\label{multi-kernel}
\end{figure}

\textbf{Local-to-local correlation.}
Specifically, CGT uses DINO \cite{caron2021emerging} as the backbone of pretrained frozen ViT.
It gets $K$-layers patch tokens $f_{q}$, $f_{s}$ and class tokens $f_{q, cls}$, $f_{s, cls}$ by inputting support images and query images and through multi-head attention.
Then, we calculate the local-to-local ($i.e.$ pixel-to-pixel) correlation between the query and support patch tokens in each layer and preserve the semantic diversity of the $M$ heads of the ViT, $i.e.$, we calculate the $M\times K$ cosine similarities of the query to support tokens and concatenate them along the new dimension:
\begin{equation}
C_{local}=\frac{(f_q)^Tf_s}{||f_q||||f_s||}\in \mathbb{R}^{MK\times h_qw_q\times h_sw_s},
\label{corr}
\end{equation}
where $h_sw_s$ and $h_qw_q$ represent the product of length and width of supports and query images, $||\cdot||$ means $l_2$ regularization.

\textbf{Local-to-global correlation.}
From the global view, we use the support mask to cut out the foreground and background regions from $f_s$.
Unlike the foreground area, which is cut off as a whole area, the background area is divided into $N$ local areas because the background may not be uniform.
To this end, we use the Voronoi-based method \cite{aurenhammer1991voronoi, zhang2022feature} to divide the background into $N$ different regions.
Then, the global features of foreground and background are obtained by mask average pooling:
\begin{equation}
\begin{aligned}
p_f&=\frac{1}{|M_s|}\sum_{i=1}^{h_sw_s}f_{s,i}M_{s,i},\\
p_{b,n}&=\frac{1}{|B_s^n|}\sum_{i=1}^{h_sw_s}f_{s,i}B_{s,i}^n,
\end{aligned} 
\label{pfb}
\end{equation}
where $M_s$ is the pseudo-mask for support images, its generation will be introduced in Section~\ref{mask}.
$B_s^n=1-M_s$ is the $n$-th background mask for the support mask. 
Similar to Eq.~\ref{corr}, the local-to-global correlation between the query and support token is calculated as follows:
\begin{equation}
\begin{aligned}
C_{global}^f&=\frac{(f_q)^Tp_f}{||f_q||||p_f||}\in \mathbb{R}^{MK\times h_qw_q\times 1},\\
C_{global, n}^b&=\frac{(f_q)^Tp_{b,n}}{||f_q||||p_{b,n}||}\in \mathbb{R}^{MK\times h_qw_q\times N},\\
\end{aligned} 
\label{cglobal}
\end{equation}
where $N=5$ is the number of the background.
We further concatenate the features to obtain the correlation token $\mathbf{C}_{i}^0\in \mathbb{R}^{(1+N+h_sw_s)\times ML}$, where $i\in[1,\cdots,h_qw_q]$ is an index over the query token and $L$ is the number of the transformer layers.
The correlation token refers to the token obtained after feeding the correlation map into the transformer.
Then following CST \cite{kang2023distilling}, it takes $\mathbf{C}_{i}^0$ and support mask $M_s$ as input and returns three types of token: foreground, background, and local correlation token through a two-layer transformer \cite{vaswani2017attention}.
Each transformer layer can be described as follows:
\begin{equation}
\begin{aligned}
\mathbf{C}_{i}^{l \prime}&=\mathrm{LN}_l(\mathrm{MHSA}_l(\mathbf{C}_{i}^l,M_{s, i})+\mathbf{C}_{i}^l),\\
\mathbf{C}_{i}^{l+1}&=\mathrm{LN}_l(\mathrm{MLP}_l(\mathbf{C}_{l}^{l \prime})+\mathbf{C}_{i}^{l \prime})\in \mathbb{R}^{C_l\times h_qw_q\times h_lw_l},
\end{aligned}
\label{cil}
\end{equation}
where $l$ means the transformer layer index, $C_l$ means its dimension, and $\mathrm{LN}_l$, $\mathrm{MHSA}_l$, $\mathrm{MLP}_l$ correspond to a multi-head self-attention (MHSA) \cite{vaswani2017attention}, a group normalization \cite{wu2018group}, and a linear layer, respectively.
Similar to related works \cite{dai2020funnel,kang2023distilling}, in each MHSA layer, the generated query is embedded into a spatial pool, and the output size changes from $h_sw_s$ to 1.
Then, we split the tensor $\mathbf{C}$ into foreground, background, and local correlation token along the second dimension, $i.e.$ $\mathbf{C}_{global}^f,\mathbf{C}_{global, n}^b,\mathbf{C}_{local}$.

\textbf{Fore-background fusion.}
After obtaining the global correlation tokens, we propose an adaptive fusion method for different global foreground and background features.
For different backgrounds, we select them with adaptive weighting, which consists of a simple linear layer.
\begin{equation}
 \mathbf{C}_{global}^b=w_1\mathbf{C}_{global, 1}^b
+ ... + w_n\mathbf{C}_{global, n}^b + \beta,
\label{cgb}
\end{equation}
where $w_n$ means the $n$-th weight of the linear layer, and $\beta$ means the bias.
Then, we fuse the merged background and foreground correlation features by a convolutional layer:
\begin{equation}
 \mathbf{C}_{global}=\textrm{Conv}_{1}(\textrm{Cat}(\mathbf{C}_{global}^f, \mathbf{C}_{global}^b)),
 \label{cg}
\end{equation}
where $\textrm{Cat}(\cdot,\cdot)$ is the concatenation operation.
This method can help the model integrate necessary support knowledge from different backgrounds and foregrounds.

\textbf{Multi-kernel information fusion.}
We use the multi-kernel information fusion mechanism after obtaining the local and global features.
Multi-kernel information fusion uses different receptive field convolution kernels to fuse the local and global correlation information, reducing the noise of different matching information due to the lack of GT masks.
We process the features by concatenating two parts of the features, utilizing convolutional kernels of different receptive fields, and helping the model learn robust knowledge:
\begin{equation}
 \mathbf{C}_i=\textrm{Conv}_{i}(\textrm{Cat}(\mathbf{C}_{local}, \mathbf{C}_{global})),
 \label{ci}
\end{equation}
where $\textrm{Conv}_i$ means the $i \times i$ convolutional operation and $i \in \{1,3,5,7\}$.
Then, we will integrate the obtained feature knowledge of different receptive fields:
\begin{equation}
 \mathbf{C}^{\prime}=\textrm{Conv}_{1}(\textrm{Cat}(\mathbf{C}_1, \mathbf{C}_3, \mathbf{C}_5, \mathbf{C}_7)) + \mathbf{C}_{local}.
 \label{cp}
\end{equation}

Finally, the final correlation token is obtained through the residual connection layer \cite{he2016deep}:
\begin{equation}
 \mathbf{C}_{fusion}=\textrm{Conv}_{3}(\mathbf{C}^{\prime}) + \mathbf{C}^{\prime}\in \mathbb{R}^{C\times h_q\times w_q}.
 \label{cfusion}
\end{equation}

\textbf{Self-distillation loss.}
We propose a self-distillation loss for our CGT to help the model generate higher-quality robust correlation diagrams in the early stage.
We average the feature dimensions for the correlation map of each layer to get the correlation map $\hat{\mathbf{C}} \in \mathbb{R}^{h_q\times w_q}$, and use the high-level correlation map to guide the low-level correlation feature map, as follows:
\begin{equation}
\label{selfdis}
 L_{distill}=\frac{1}{L}\sum_{l=1}^{L}\sum_{i=1}^{h_lw_l}\zeta_l(\hat{\mathbf{C}}_{local,i}^{l+1})\cdot\log\frac{\zeta_l(\hat{\mathbf{C}}_{local,i}^{l+1})}{\hat{\mathbf{C}}_{local,i}^l},
\end{equation}
where $\zeta_l(\cdot)$ is the resize function of the $l$-th layer, $L$ is the number of the layers and $h_lw_l$ is the the product of length and width of the $l$-th layer.
The guidance of the high-level correlation graph to the low-level feature graph helps the model retain the fine-grained segmentation quality, reduces the impact of noise, and does not discard the context information \cite{peng2023hierarchical}, which can help learn robust correlations.

\subsection{Class-Guided Module}
\label{sec:cgm}
The knowledge learned by the model from the correlation between support and query is limited, especially in the case of imprecise support masks in the WS-FSS scenarios.
To further assist the model in filtering potential noise in correlation features, we use additional category information from the perspective of category semantics to help the model locate more valuable correlation information.
Pre-trained CLIP \cite{radford2021learning} has been proven to generate relatively coarse CAM based on category information by using Grad-CAM \cite{selvaraju2017grad,wang2023iterative}.
After large-scale pre-training, CLIP already has powerful zero-shot learning capabilities.
Even without seeing specific supervision labels during training, CLIP is able to understand and generate output for tasks for which it was not explicitly trained \cite{zhou2023zegclip,jiao2023learning}.
This paper utilizes this to construct the CGM that helps the model roughly locate the approximate positions of the objects that need to be segmented.

To simplify our method, this paper will not discuss obtaining more accurate masks for CLIP. 
Instead, we will choose a simple mask generation method and discuss utilizing the generated coarse masks.
CGM can also be seen as a simple zero-shot method, and it does not rely on various complex cue engineering and other zero-shot models but can still achieve satisfactory performance.

We first input the query image and its category prompt ``a photo of [class]", where class represents its corresponding category $c$, into the pre-trained CLIP and then use Grad-CAM to obtain a coarse attention $\mathbf{A}_c$.
Next, we multiply $\mathbf{A}_c$ by the obtained correlation token $\mathbf{C}_{fusion}$ and use $\mathbcal{F}_{CGM}$ to refine the attention:
\begin{equation}
 \mathbf{A}_r=\mathbcal{F}_{CGM}(\mathbf{C}_{fusion}\otimes\zeta(\mathbf{A}_c))\in\mathbb{R}^{h_q\times w_q},
 \label{ar}
\end{equation}
where $\mathbcal{F}_{CGM}$ consists of two convolutional layers and a sigmoid function, $\otimes$ is the Hadamard product and $\zeta(\cdot)$ is the resize function.
Finally, we combine the features with the attention $\mathbf{A}_r$ to obtain filtered correlation features that discard irrelevant background information:
\begin{equation}
 \tilde{\mathbf{C}}=(\mathbf{C}_{fusion}\otimes \mathbf{A}_r)\oplus \mathbf{C}_{fusion},
 \label{cc}
\end{equation}
where $\oplus$ stands for the element-wise sum.
Combined with backpropagation, the parameters of $\mathbcal{F}_{CGM}$ in CGM are updated.
Therefore, the refined $\mathbf{A}_r$ can focus more on the objects that need to be segmented based on the $\mathbf{A}_c$ generated by CLIP.
\replaced{Through the coarse-to-fine training strategy, when the model encounters unfamiliar categories, even without the precise support of mask supervision, it can combine the powerful zero-shot capability of CLIP to capture the approximate location of the segmented object.}{Through the coarse-to-fine training strategy, the model can capture the approximate position of segmented objects even without accurate support for mask supervision when encountering unfamiliar categories. }

\subsection{Embedding-Guided Module}
\label{sec:egm}
Towards the goal of reducing the potential information loss of the model in correlation-enhanced learning under weakly-supervised segmentation settings, we suggest embedding the original appearance of each layer obtained from support and query feature maps into the decoder for further aggregation to implicitly guide the model in utilizing the learned robust support-query matching information.

First, add the features of each layer and project them:
\begin{equation}
\begin{aligned}
\label{sum}
F_s &= \mathbcal{F}_{Proj.}(\sum_{k=1}^K f_s^k),\\
F_q &= \mathbcal{F}_{Proj.}(\sum_{k=1}^K f_q^k),
\end{aligned}
\end{equation}
where $\mathbcal{F}_{Proj.}$ denotes $1\times 1$ convolution and $K$ means the layer number of the backbone ViT.
Then they are concatenated to the similarity feature $\tilde{\mathbf{C}}$, and the final prediction mask $\tilde{M}_q$ is obtained through the EGM composed of two layers of transformers \cite{liu2021swin} and a segmentation header:
\begin{equation}
 \tilde{M}_q = EGM(Cat(\tilde{\mathbf{C}}, F_s, F_q)).
 \label{mq}
\end{equation}

The original appearance information implicitly helps the model reduce information loss in learning robust correlation.
Meanwhile, the appearance embedding information is an effective guide for filtering noise in matching scores \cite{hosni2012fast,sun2018pwc,hong2022cost}, while self-supervised pre-trained ViT can provide an efficient multi-layer feature.
It helps the model learn the relevant information obtained in the presence of certain mismatches through embedding guidance at each layer.

\subsection{Training Objective}\label{mask}
\textbf{Pseudo-mask.}
As demonstrated in previous studies \cite{amir2021deep,caron2021emerging,kang2023distilling}, query key attention maps can capture semantically significant foreground objects.
Inspired by this, we generate a pseudo-GT mask for dynamic queries and image support by calculating the cross attention of the last ViT layer:
\begin{equation}
\begin{aligned}
M_{s,i}^m&=\frac{(f_{s,i}^m)^Tf_{q,cls}^m}{||f_{s,i}^m||||f_{q,cls}^m||}\in \mathbb{R}^{h_sw_s\times 1},\\
M_{q,i}^m&=\frac{(f_{q,i}^m)^Tf_{s,cls}^m}{||f_{q,i}^m||||f_{s,cls}^m||}\in \mathbb{R}^{h_qw_q\times 1},
\end{aligned}
\label{msmq}
\end{equation}
where $f_{q,cls}^m$, $f_{s,cls}^m$ means $m$-th head query or support class token.
Meanwhile, we use the Pixel-Adaptive Refinement (PAR) module \cite{ru2022learning} to generate pseudo-masks based on the relationship information between various pixels within the image, generating more accurate supervision information:
\begin{equation}
\begin{aligned}
M_{s,i}&=\mathds{1}(PAR(\zeta(\frac1M\sum_{m=1}^MM_{s,i}^m)>\alpha)),\\
M_{q,i}&=\mathds{1}(PAR(\zeta(\frac1M\sum_{m=1}^MM_{q,i}^m)>\alpha)),
\end{aligned}
\label{msimqi}
\end{equation}
where $\alpha=0.4$ is the prediction threshold and $\mathds{1}(\cdot)$ is the indicator function.
Unlike existing WS-FSS methods \cite{zhang2022weakly}, our mask generation module also applies to unseen categories without additional training stages.

\textbf{Training loss.}
There are two parts of training loss: segmentation loss and self-distillation loss.
The segmentation loss $L_{seg}$ is calculated by the final prediction $\tilde{M}_q$ and $M_s$ using the cross-entropy function.
The self-distillation loss is obtained from Eq.~\ref{selfdis}.
The final loss is:
\begin{equation}
 L = L_{seg} + \lambda_{distill}L_{distill},
 \label{loss}
\end{equation}
where $\lambda_{distill}$ is the balance parameter set to 0.5. The whole training process for CORENet is summarized in Algorithm \ref{alg}.

\begin{algorithm}[!ht]
    \caption{Training Process for CORENet.}
    \label{alg}
    \SetAlgoLined
    \KwIn{A training set $D_{train}$ and a training category set $C_{train}$.}
    \KwOut{The final trained model $\phi$.}
    \For{each episode ($S$, $Q$) $\in$ $D_{train}$ and category $c$ $\in$ $C_{train}$}{
      Extract features by pretrained DINO ViT.\\
      Generate the pseudo-mask using Eq.\ref{msmq} and \ref{msimqi}.\\
      \textcolor{gray}{\textit{\# Correlation-Guided Transformer}}\\
      Compute the local-to-local and local-to-global correlation using Eq.~\ref{corr},~\ref{pfb} and \ref{cglobal}.\\
      Obtain foreground, background, and local correlation tokens using Eq.~\ref{cil}.\\
      Obtain the final correlation token $\mathbf{C}_{fusion}$ using Eq.~\ref{cgb},~\ref{cg},~\ref{ci},~\ref{cp} and~\ref{cfusion}.\\
      Compute the self-distillation loss $L_{distill}$ as in Eq.~\ref{selfdis}.\\
      \textcolor{gray}{\textit{\# Class-Guided Module}}\\
      Obtain filtered correlation feature $\tilde{\mathbf{C}}$ using Eq.~\ref{ar} and~\ref{cc}.\\
      \textcolor{gray}{\textit{\# Embedding-Guided Module}}\\
      Predict the query mask $\tilde{M}_q$ using Eq.~\ref{sum} and~\ref{mq}.\\
      Compute the final loss $L$ as in Eq.~\ref{loss}.\\
      Compute gradients and optimize via SGD.
    }
    \textbf{Return} the final trained model $\phi$.
\end{algorithm}

\section{Experiments}
\label{experiments}
In this section, we evaluate the proposed method, compare it with recent state-of-the-art, and provide in-depth analyses of the results of the ablation study.
\subsection{Experimental Settings}
\textbf{Datasets.}
To evaluate our method, experiments are conducted on two commonly used few-shot segmentation datasets, PASCAL-5$^i$ and COCO-20$^i$. 
PASCAL-5$^i$ is created according to PASCAL VOC 2012 \cite{everingham2010pascal} with additional notes of SBD \cite{hariharan2014simultaneous}. 
A total of 20 classes in the dataset are evenly divided into four folds $i$ $\in$ \{0, 1, 2, 3\}, and each fold contains five classes.
COCO-20$^i$ is proposed by \cite{nguyen2019feature} and is based on MSCOCO \cite{lin2014microsoft}. 
Similar to PASCAL-5$^i$, the 80 classes in COCO-20$^i$ are divided into four folds, and each fold contains 20 classes.

\textbf{Evaluation metrics.}
We use union average intersection (mIoU) as our evaluation indicators.
The mood indicator averages the IoU values of all classes in the fold: $\text{mIoU}=\frac{1}{C}\sum^{C}_{c=1}\text{IoU}_c$, where $C$ is the number of classes in the target fold and $\text{IoU}_c$ is the intersection on the union of class $c$.
Because mIoU better reflects the generalization ability and prediction quality of the model, we mainly focus on mIoU in our experiments.

\textbf{Implementation details.}
To compare with previous works based on ResNet50 \cite{min2021hypercorrelation,kang2022integrative,yang2023mianet}, we use the ViT-small backbone \cite{dosovitskiy2020image}.
The feature extraction backbone network conducts self-monitoring and pre-training on ImageNet 1K \cite{russakovsky2015imagenet} through DINO \cite{caron2021emerging}.
Following the CST \cite{kang2023distilling}, the reason for choosing this ViT is that its training data size and the number of model parameters are similar to ResNet50 \cite{he2016deep}.
It is also trained on ImageNet 1K but uses class labels as supervision.
The backbone of CLIP is ResNet101.
However, our framework based on DINO and CLIP can easily replace the backbone network with a foundation model such as ViT-G/14 \cite{zhai2022scaling} with a huge parameter amount (2.5B), which distinguishes our method from existing methods.
As in the previous works \cite{min2021hypercorrelation,kang2022integrative,kang2023distilling}, the backbone is frozen during training.
The learning rate is initialized to 0.0005, the batch size is 16, and the additional layer is trained using Adam \cite{kingma2014adam}.
The loss balance parameter $\lambda_{distill}$ is set to 0.5, and the number of backgrounds $N$ is set to 5.
Consistent with CST \cite{kang2023distilling}, our CORENet uses a 1-way 1-shot segment for training and any $N$-way $K$-shot inference.

\begin{table*}[t]
\caption{Performance of PASCAL-5$^i$ \cite{everingham2010pascal} in mIoU.
The superscript $*$ indicates that the model is trained on the pseudo-mask generated by CST \cite{kang2023distilling}.
\textbf{Bold} numbers indicate the best performance, and \underline{underlined} numbers indicate the second best.}
\centering
\scalebox{0.7}{
\begin{tblr}{
  cells = {c},
  cell{1}{2} = {c=5}{},
  cell{1}{7} = {c=5}{},
  vline{2-3} = {1}{},
  vline{7-8} = {1}{},
  vline{2,7} = {2-8}{},
  hline{1,9} = {-}{0.08em},
  hline{2-3,7-8} = {-}{0.05em},
}
 & 1-shot &  &   &   &   & 5-shot &  &    &   &   \\
Methods & $5^0$  & $5^1$  & $5^2$  & $5^3$  & mean & $5^0$  & $5^1$  & $5^2$  & $5^3$  & mean \\
HSNet$^*$ \cite{min2021hypercorrelation}  & 47.6&	\underline{45.4}	&41.0&	\underline{37.0}	&42.8&  48.0	&\underline{46.1}	&41.6&	\underline{37.3}	&43.3\\
ASNet$^*$ \cite{kang2022integrative}& \underline{49.0}&	44.6	&\underline{43.8}	&35.2&	\underline{43.2} &\underline{50.1}&	45.6	&45.0	&35.8	&\underline{44.1}  \\
MIANet$^*$ \cite{yang2023mianet}& 44.9	&34.3	&41.2	&35.9	&39.1& 46.4	&45.1	&\underline{46.5}	&36.4	&43.6\\
CST \cite{kang2023distilling}  & 48.2 &	45.1	&42.4	&34.6	&42.5 & 49.5 & 45.5 & 42.8 & 35.1  & 43.2    \\
CORENet (Ours) & \textbf{50.3}	&\textbf{51.6}	&\textbf{47.6}	&\textbf{39.4}	&\textbf{47.2} & \textbf{50.7}	&\textbf{51.8}&	\textbf{47.8}&	\textbf{39.6}&	\textbf{47.5}
\end{tblr}}
\label{voc}
\end{table*}

\begin{table*}[t]
\caption{Performance of COCO-20$^i$ \cite{nguyen2019feature} in mIoU.
The superscript $*$ indicates that the model is trained on the pseudo-mask generated by CST \cite{kang2023distilling}.
\textbf{Bold} numbers indicate the best performance, and \underline{underlined} numbers indicate the second best.}
\centering
\scalebox{0.7}{
\begin{tblr}{
  cells = {c},
  cell{1}{2} = {c=5}{},
  cell{1}{7} = {c=5}{},
  vline{2-3} = {1}{},
  vline{7-8} = {1}{},
  vline{2,7} = {2-6}{},
  hline{1,7} = {-}{0.08em},
  hline{2-3,6} = {-}{0.05em},
}
               & 1-shot &     &     &     &              & 5-shot &          &     &      &        \\
Methods & $20^0$  &$20^1$  & $20^2$  & $20^3$  & mean  & $20^0$  & $20^1$  & $20^2$  & $20^3$  & mean  \\
HSNet$^*$ \cite{min2021hypercorrelation}  & 19.9	&\underline{22.5}	&22.1	&\underline{23.0}	&21.9 &21.0	&\underline{24.2}	&\textbf{22.7}	&\underline{23.8}	&\underline{22.9}     \\
MIANet$^*$ \cite{yang2023mianet}& 20.5	&\textbf{22.8}	&21.6	&22.3	&21.8 &\underline{21.5}	&23.2	&21.8	&22.5	&22.3\\
CST \cite{kang2023distilling}             & \underline{21.0}	&21.9	&\textbf{22.4}	&22.5	&\underline{22.0}  & 21.3    & 22.1    & \textbf{22.7} & 22.6 &  22.1    \\
CORENet (Ours) & \textbf{22.1}	&\textbf{22.8}&	\underline{22.3}	&\textbf{23.4}	&\textbf{22.7}& \textbf{22.3}&	\textbf{24.7}&	\underline{22.6}&	\textbf{24.0}	&\textbf{23.4}
\end{tblr}}
\label{coco}
\end{table*}

\begin{figure}[htbp]
\centering
\includegraphics[scale=0.16]{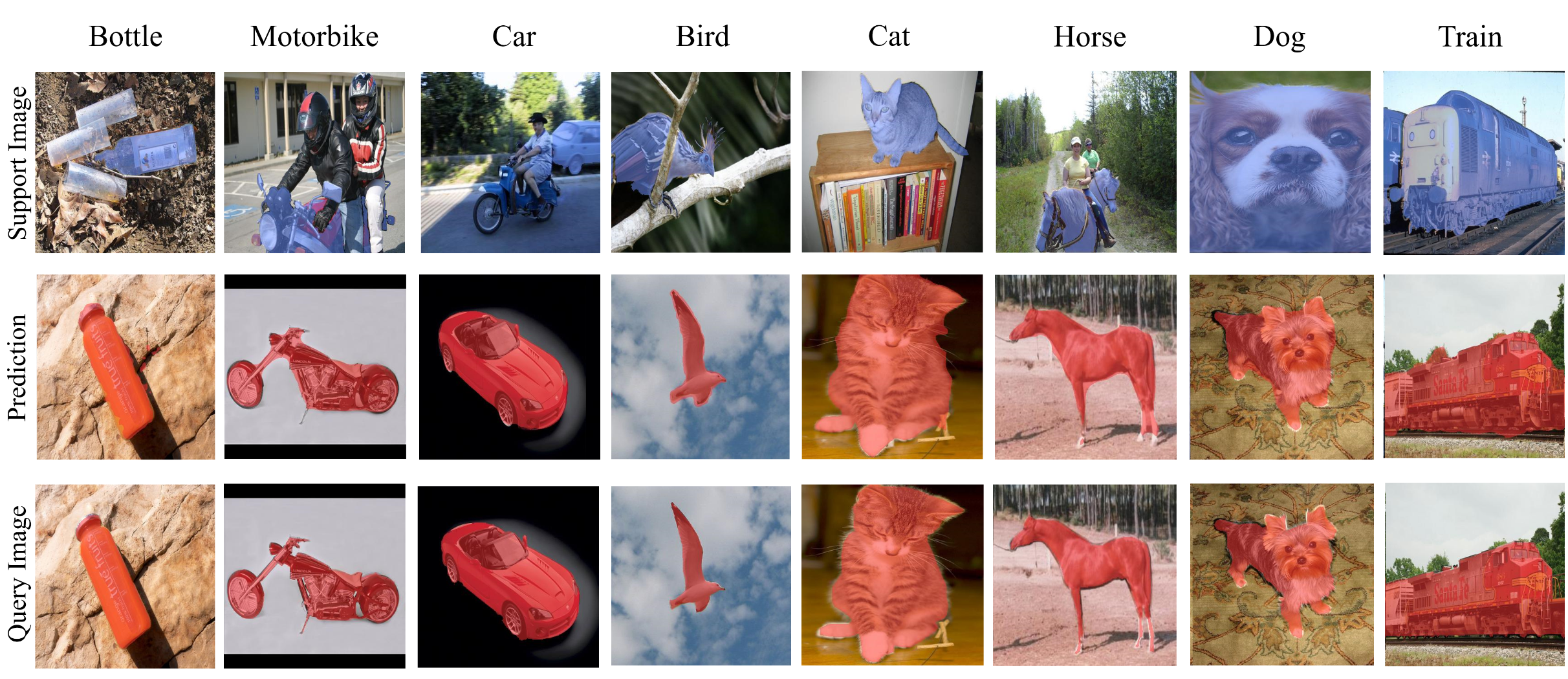}
\caption{Qualitative results of our CORENet on PASCAL-5$^i$ and COCO-20$^i$ benchmarks. 
Zoom in for details.}
\label{visbig}
\end{figure}

\subsection{Comparison with State-of-the-Arts.}
Due to the lack of supervision masks, the existing FSS model cannot be directly migrated to the WS-FSS scenario.
To better compare existing FSS methods, we combine them with the mask generation method in CST \cite{kang2023distilling} to generate supervised information and guide the model in predicting the final query mask.
We label them with the superscript $*$.

\textbf{PASCAL-5$^i$.}
Table \ref{voc} compares mIoU performance between our method and existing representative models.
From this, it can be seen that:
(i) CORENet achieved state-of-the-art performance in both 1-shot and 5-shot settings.
Compared to the recent FSS model MIANet \cite{yang2023mianet} and the weakly-supervised classification \& segmentation model CST \cite{kang2023distilling}, we have improved by 8.1\%, 4.7\% (1-shot), and 3.9\%, 4.3\% (5-shot), respectively.
(ii) MIANet has performed poorly in certain situations, which holds the previous
state-of-the-art results of FSS.
This is because in WS-FSS scenarios, the generated mask contains noise, and excessive dependence on correlated features with noise can lead to a decrease in model performance.
This also confirms that the method proposed in this paper can effectively handle scenarios with mask noise in weakly-supervised few-shot segmentation.

\textbf{COCO-20$^i$.}
COCO-20$^i$ is a more challenging dataset containing multiple objects and more significant variance.
Table \ref{coco} shows the performance comparison of mIoU.
Overall, the mean mIoU of MIANet in the 1-shot and 5-shot settings surpasses all previous methods.
Under the 1-shot setting, our CORENet exceeded MIANet and CST by 0.9\% and 0.7\%.
This proves the superiority of our method despite the challenging scenarios.

\textbf{Qualitative results.}
Fig.~\ref{visbig} reports quantitative results from CORENet and baseline models based on PASCAL-5$^i$ and COCO-20$^i$ benchmark tests.
We can see that CORENet performs well in capturing object details.
For example, more subtle details are retained in segmenting dogs and cars.

\subsection{Ablation Study}
We conducted extensive ablation studies of PASCAL-5$^i$ to verify the effectiveness of the critical modules (CGT, CGM, and EGM) we proposed.
In addition, we provide experimental details and additional experiments in the supplementary materials.

\textbf{Components analysis.}
Our CORENet consists of four key components: Correlation-Guided Transformer (CGT), Class-Guided Module (CGM), Emb-edding-Guided Module (EGM), and Pixel-Adaptive Refinement (PAR).
Table \ref{abs} shows our validation of the effectiveness of each component.
PAR can further reduce imprecise noise in masks and achieve a performance improvement of 2.1\% in 1-shot by fully utilizing the information of surrounding pixels.
EGM is an essential component of our model, which increases mIoU by 1.1\% in 1-shot.
CGT and CGM are also indispensable.
By combining all three modules, CORENet achieves state-of-the-art performance.

\begin{table}[t]
\caption{Ablation studies of main model components.
\textbf{Bold} numbers indicate the best performance.}
\centering
\scalebox{0.8}{
\begin{tabular}{ccccc|cc}
\toprule
Baseline & CGT & CGM & EGM & PAR & 1-shot & 5-shot \\
\midrule
\checkmark &   &  &     &     & 42.5      & 43.2  \\
  \checkmark &  \checkmark &  &     &     & 43.1      & 43.9  \\
 \checkmark &\checkmark   &  \checkmark &     &     & 44.0   &   44.7   \\
 \checkmark &\checkmark   &  \checkmark &  \checkmark   &     & 45.1   & 45.8    \\
 \checkmark &\checkmark   &  \checkmark &  \checkmark   &  \checkmark   & \textbf{47.2}  & \textbf{47.5} \\
\bottomrule
\end{tabular}}
\label{abs}
\end{table}

\begin{table*}[t]
\begin{floatrow}
\capbtabbox{
\renewcommand{\arraystretch}{1.2}
\scalebox{0.86}{
\begin{tabular}{cccc|cc}
\toprule
FBC & FBF & SIF & MIF & 1-shot & 5-shot \\
\midrule
  \checkmark &  & \checkmark    &     & 46.6      & 46.9  \\
  \checkmark &  &   &    \checkmark   & 46.9   & 47.1     \\
  &\checkmark    &   \checkmark  &     & 46.7   & 47.2     \\
 &\checkmark    &   &  \checkmark   & \textbf{47.2}  & \textbf{47.5} \\
\bottomrule
\end{tabular}}
}{
\caption{Ablation studies of main components in CGT.
\textbf{Bold} numbers indicate the best performance.}
\label{cgt}
}
\capbtabbox{
\renewcommand{\arraystretch}{1.2}
\scalebox{0.84}{
\begin{tabular}{c|c}
\toprule
Background regions & 1-shot  \\
\midrule
  $N=4$ & 47.0 \\ 
  $N=5$ & \textbf{47.2} \\ 
  $N=6$ & 47.1 \\
  $N=7$ & \textbf{47.2} \\
\bottomrule
\end{tabular}}
}{
\caption{Analysis of background regions in CGT.}
\label{background}
}
\end{floatrow}
\end{table*}

\textbf{Main components in CGT.}
CGT constructs local-to-global correlations to help the model fully understand matching information.
Table \ref{cgt} shows the impact of each element in CGT on model performance.
``FBC” means the fore-background concatenation, ``FBF" means the fore-background fusion, ``SIF” denotes the single-kernel information fusion, and ``MIF” denotes the multi-kernel information fusion.
We can see that the fusion using adaptive weights is 0.3\% better than the method of directly concatenating background features along the channel.
In addition, by establishing path information between different receptive fields, the proposed multi-kernel information fusion method is compared to the single-kernel fusion method (using only 3 × 3 convolutions), which can further reduce the impact of mismatches on the model and achieve better performance.
Through the proposed fusion mechanism, CGT can help the model learn more robust correlation with more information while support mask inaccuracies.

\begin{table*}[t]
\begin{floatrow}
\capbtabbox{
\renewcommand{\arraystretch}{1.1}
\scalebox{0.9}{
\begin{tabular}{cc|c}
\toprule
CLIP & Refinement & 1-shot \\
\midrule
  &     & 46.5  \\
\checkmark  &     & 46.8  \\
  \checkmark &  \checkmark  & \textbf{47.2}  \\
\bottomrule
\end{tabular}}
}{
\caption{\replaced{Ablation studies of CGM.}{Analysis of refinement in CGM.}}
\label{clip}
}
\capbtabbox{
\renewcommand{\arraystretch}{1.1}
\scalebox{0.92}{
\begin{tabular}{c|c}
\toprule
Dimension  & 1-shot  \\
\midrule
  32 & 46.8 \\ 
  64 & \textbf{47.2} \\
  128 &  46.9 \\
 384  &  47.1  \\
\bottomrule
\end{tabular}}
}{
\caption{Analysis of projection dimension in EGM.}
\label{dim}
}
\end{floatrow}
\end{table*}

\textbf{Number of background regions for CGT.}
In Section~\ref{sec:cgt}, we mentioned that the Voronoi-based method \cite{aurenhammer1991voronoi,zhang2022feature} helps the model learn complex background correlation knowledge by dividing different background regions.
We further discuss the impact of this region on the final results of the model, as shown in Table \ref{background}.
It can be seen from the results in the table that the model results are relatively robust for different numbers of regions.
This shows that our CGT can perform robust correlation modeling for different complex background knowledge, which can help the model still learn valuable correlation knowledge when facing the generated imprecise masks.

\textbf{Refinement of CGM.}
In Section~\ref{sec:cgm}, we mentioned using pre-trained CLIP \cite{radford2021learning} assisted models for segmentation by constructing CGM.
We further discuss the necessity of pre-trained CLIP and the proposed enhancement module, as shown in Table \ref{clip}.
\replaced{When no additional components are added, the model directly feeds the correlation features obtained through CGT into the EGM module.
For the different approaches to fusing attention map of CLIP, we compare the approach of simply fusing it into the original correlation features (denoted as “CLIP”) with the approach of fusing it by designing additional learnable refinement modules (denoted as “Refinement”).
Due to the zero-shot capability of CLIP, the performance of the model can be improved to a certain extent when only CLIP is used to weight features.
The experimental results show that after the refinement module, the model can better combine the knowledge provided by CLIP to help the model filter out irrelevant background areas and achieve the best results.}
{As shown in the first line, if only CLIP is used for segmentation, the segmentation results are very limited.
However, ideal results can be obtained if the results of CLIP segmentation are directly combined with the correlation features obtained by our proposed CGT without enhancement modules.
Finally, a simple enhancement module helps the model filter out irrelevant background regions and achieve the best results.}

\begin{figure}[t]
\centering
\begin{subfigure}{0.3\linewidth}
    \centering
    \includegraphics[width=1\linewidth]{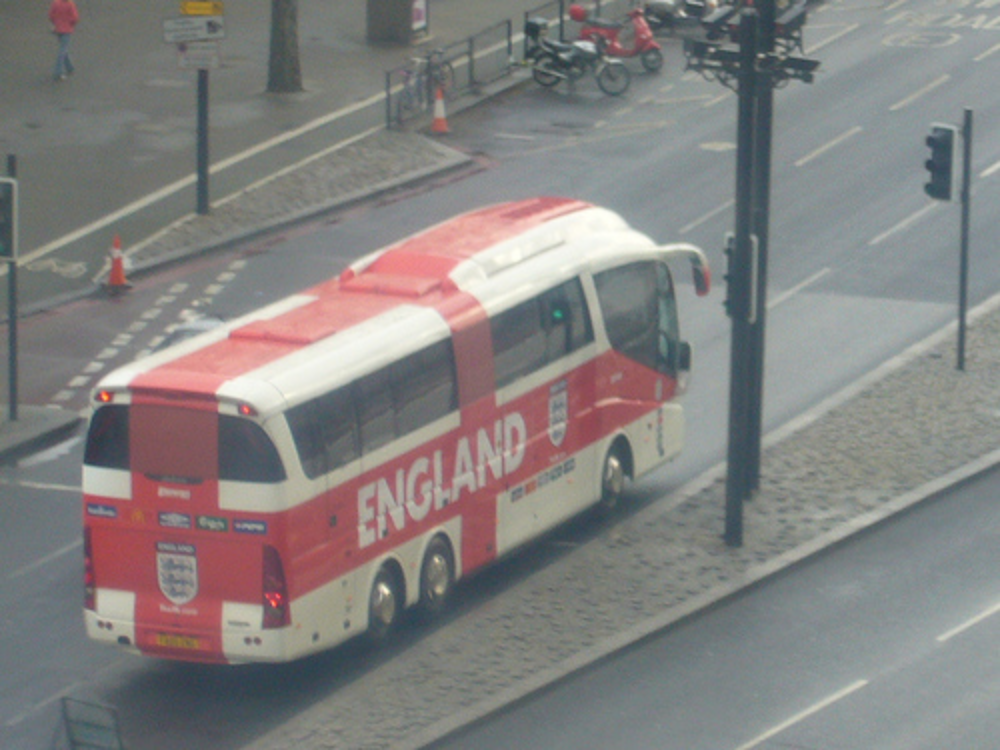}
    \caption{Original image}
    \label{chutian1}
\end{subfigure}
\centering
\begin{subfigure}{0.3\linewidth}
    \centering
    \includegraphics[width=1\linewidth]{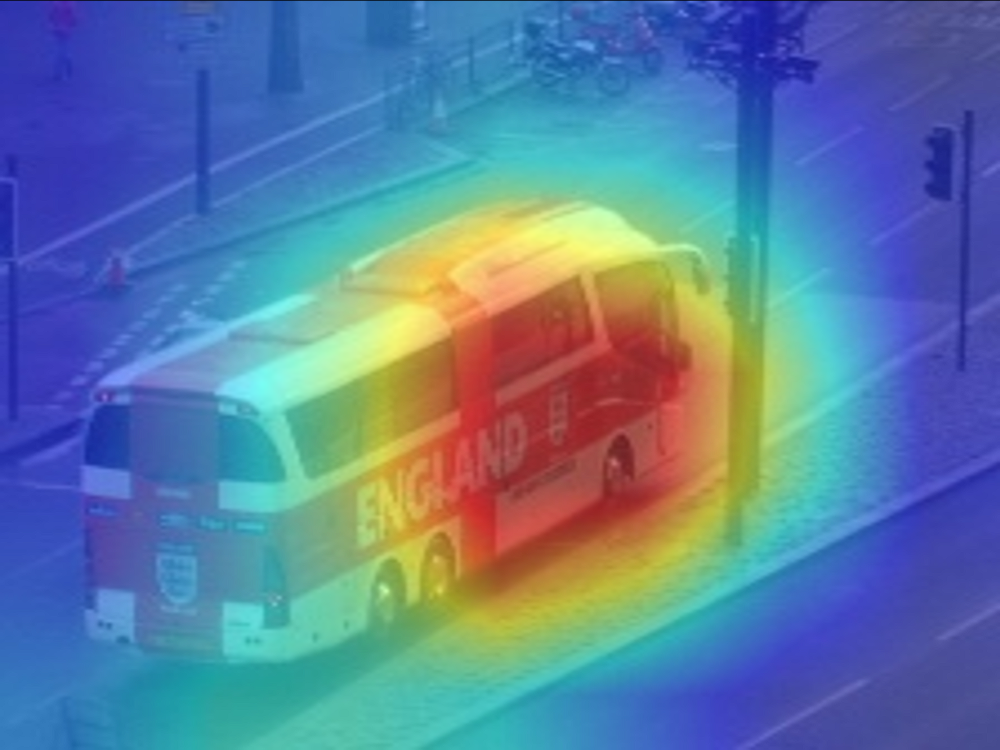}
    \caption{ResNet50}
    \label{chutian2}
\end{subfigure}
\centering
\begin{subfigure}{0.3\linewidth}
    \centering
    \includegraphics[width=1\linewidth]{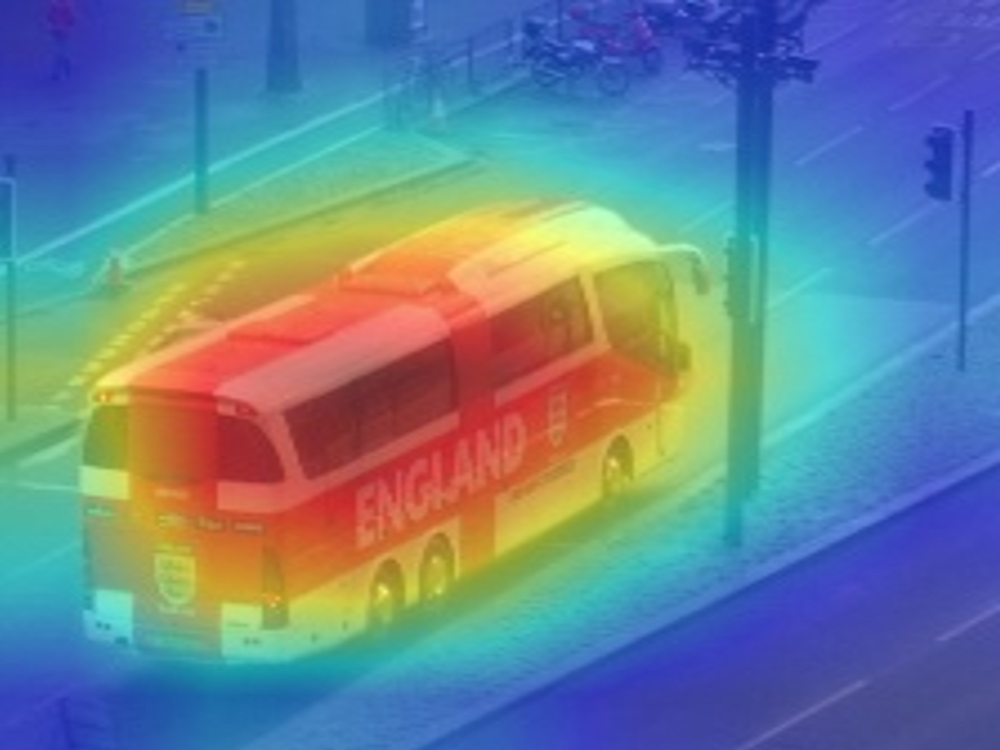}
    \caption{ResNet101}
    \label{chutian3}
\end{subfigure}
\caption{Visualization of GradCAM obtained from different backbones of CLIP in CGM.}
\label{da_chutian}
\end{figure}

\textbf{Differenct backbone of CLIP in CGM.}
We use the GradCAM of the pre-trained CLIP in the CGM module to generate initialization attention maps.
As shown in Fig. \ref{da_chutian}, we visualized the thermal maps obtained by CLIP for different backbones.
When using a deeper network as the backbone of CLIP, the initial attention map obtained is visually better.
A better initial attention map can help the model focus on more critical correlation information.
Therefore, we chose ResNet101 \cite{he2016deep} as the backbone of our CGM module.

\textbf{Projection dimension of EGM.}
In Section~\ref{sec:egm}, we mentioned projecting the original feature representation onto a particular dimension and concatenating it into the enhanced correlation features to reduce potential information loss during enhancement.
We conducted experiments on different projection dimensions, and the results are shown in Table \ref{dim}.
Different projection dimensions have little impact on the experimental results, and the best effect is achieved when the dimension is 64.

\begin{table*}[t]
\begin{floatrow}
\capbtabbox{
\renewcommand{\arraystretch}{1.1}
\scalebox{0.85}{
\begin{tabular}{c|c}
\toprule
Dimension  & 1-shot  \\
\midrule
  Concatenation & 46.9 \\ 
  Sum & \textbf{47.2} \\
\bottomrule
\end{tabular}}
}{
\caption{Analysis of differences in projection operations of EGM.}
\label{sum2}
}
\capbtabbox{
\renewcommand{\arraystretch}{1.1}
\scalebox{0.8}{
\begin{tabular}{c|cc}
\toprule
Method  & 1-shot & 5-shot \\
\midrule
  Pixel-level meta-learner \cite{lee2022pixel}& 42.4 & 45.5 \\ 
  CORENet (Ours)& \textbf{47.2}&\textbf{47.5} \\
\bottomrule
\end{tabular}}
}{
\caption{Performance differences with related methods \cite{lee2022pixel} in PASCAL-5$^i$.}
\label{samework}
}
\end{floatrow}
\end{table*}

\textbf{Differences in projection operations of EGM.}
In Eq.~\ref{sum}, we mentioned the operation of adding all features during the feature projection process.
A feasible alternative is to concatenate all features and feed them into a dimensionally reduced convolutional layer.
The features obtained in this way are of the same size as those obtained by direct addition, where we note this scheme as "concatenation".
The comparison results of the two schemes are shown in Table~\ref{sum2}. It can be seen that directly adding features can obtain better results than concatenation without the need for additional convolution operations.

\textbf{Comparison with existing similar work.}
It is worth noting that the recent related work \cite{lee2022pixel} also considers a similar problem setting.
Different from \cite{lee2022pixel}, our method focuses more on exploring the performance capabilities of the foundation model in WS-FSS.
To further demonstrate the difference between the two methods, we compare the differences between the two methods on PASCAL-5$^i$, and the results are shown in Table~\ref{samework}.
It can be seen from the results that our method can help the model perform better in the WS-FSS scenario due to its strong generalization ability based on the foundation model and the robustness of the proposed method.

\begin{figure}[tb]
\centering
\includegraphics[scale=0.15]{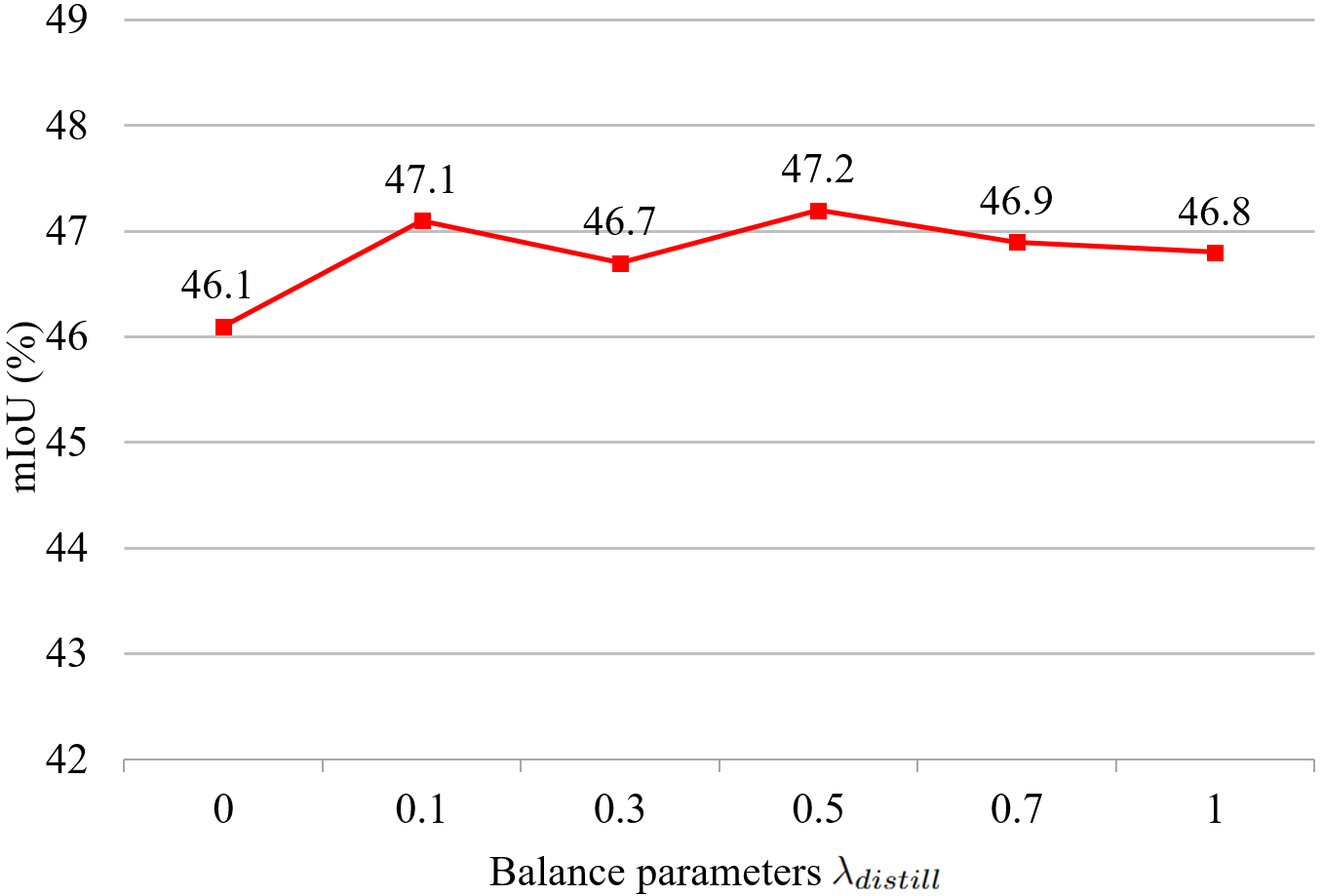}
\caption{Results with loss balance parameter $\lambda_{distill}$ on the 1-shot setting.}
\label{sensi}
\end{figure}

\textbf{Parameter sensitivity.}
For our CGT, we have designed a self-distillation loss aid model to generate higher-quality correlation maps.
We conducted sensitivity experiments on different loss balance parameters $\lambda_{distill}$, as shown in Fig.~\ref{sensi}.
Under different $\lambda_{distill}$, the mIoU variation of the model is relatively robust and reaches its optimal value at 0.5.
However, without using self-distillation loss, $i.e.$ $\lambda_{distill}=0$, the model performance decreases by 1.1\%, further proving the advantage of our proposed loss.

\subsection{Discussions}
\textbf{Discussions about related settings.}
\label{sec:Discussion.}
We demonstrate the differences between WS-FSS and related work settings in Table~\ref{diff}.
Compared to WS-FSS, few-shot segmentation methods \cite{hong2022cost, yang2023mianet, wang2019panet, min2021hypercorrelation, yang2021mining, siam2019amp, liu2020crnet} rely more on precise ground-truth masks and learn to support and query the correlation of images based on this.
Due to differences in application scenarios, weakly-supervised segmentation methods \cite{li2016weakly, bilen2016weakly, ru2022learning} cannot segment new classes that have not been seen before.
Most relevant to WS-FSS, the weakly-supervised few-shot classification \& segmentation method \cite{kang2023distilling} not only performs weakly-supervised segmentation on query images containing unseen categories but also allows the model to output whether they belong to the same category as the support images.
However, the category supervision information it provides is whether the query image is in the same category as the support image. 
It does not provide specific category information to assist the model in FSS.

\begin{table*}[t]
\renewcommand{\arraystretch}{1.1}
\centering
\scalebox{0.61}{
\begin{tabular}{ll}
\toprule
\textbf{Related task settings}  & \textbf{Difference}  \\
\midrule
  Few-shot semantic segmentation \cite{shaban2017one} & Dependence on accurate ground-truth mask.    \\
\midrule
  Weakly-supervised semantic segmentation \cite{bilen2016weakly} & The model can only segment seen categories.    \\
\midrule
 Weakly-supervised few-shot classification \& segmentation \cite{kang2023distilling} & \makecell[l]{The provided category information is whether the two\\  images belong to the same category (0/1 label) and does \\not provide specific category assistance for segmentation.}    \\

\bottomrule
\end{tabular}}
\caption{The difference between related work and weakly-supervised few-shot segmentation.}
 \label{diff}
\end{table*}

\textbf{Feature work.}
Compared to state-of-the-art methods on relatively simple datasets, our method has succeeded considerably. 
However, model performance can continue to improve when faced with more complex datasets (such as COCO-20$^i$).
We will explore in the future how to better learn correlations between more complex images.
On the other hand, our random division of background region in CGT also considers this problem. 
More complex correlations are learned by setting a more significant background number $N$, and the problem becomes part of parameter selection.
In the future, we will have a more in-depth discussion on this issue to help the model learn more robust correlation knowledge without GT masks.

\section{Conclusion}
\label{conclusion}
This paper proposes a framework to address the issue of requiring precise masks for existing FSS tasks, which address weakly-supervised few-shot segmentation tasks with only category information.
To better mine the robust correlation between support queries, this paper proposes that CGT calculate similarity information from global and local perspectives.
Then, from the perspective of category semantics, we designed CGM to help the model roughly locate targets using the pre-trained CLIP.
In addition, the EGM module was designed to implicitly guide the model in filtering noise in correlation from the perspective of appearance embedding.
Extensive experiments have shown that our CORENet has achieved state-of-the-art results in our weakly-supervised few-shot segmentation tasks.

{
    \bibliographystyle{elsarticle-num}
    \bibliography{main}
}
\end{document}